\documentclass[conference]{IEEEtran}
\IEEEoverridecommandlockouts
\usepackage{cite}
\usepackage{amsmath,amssymb,amsfonts}
\usepackage{algorithmic}
\usepackage{graphicx}
\usepackage{booktabs} % For formal tables
\usepackage{makecell}
\usepackage{textcomp}
\usepackage{xcolor}
\usepackage{url}
\usepackage{version}

\usepackage[breaklinks=true]{hyperref}
\usepackage{breakcites}

\def\BibTeX{{\rm B\kern-.05em{\sc i\kern-.025em b}\kern-.08em
    T\kern-.1667em\lower.7ex\hbox{E}\kern-.125emX}}
\begin{document}

\title{A Methodology for Ethics-by-Design AI Systems: Dealing with Human Value
Conflicts}
%A Methodology for building Ethics-by-Design AI Systems: Examples of Intelligent System Applications dealing with Human Value Conflicts}

\author{\IEEEauthorblockN{Fabrice Muhlenbach}
\IEEEauthorblockA{\textit{Universit\'{e} de Lyon, UJM-Saint-Etienne, CNRS, Laboratoire Hubert Curien, UMR 5516}\\
18 rue du Professeur Beno\^{\i}t Lauras, F-42023 Saint Etienne, FRANCE \\
E-mail: \texttt{fabrice.muhlenbach@univ-st-etienne.fr} \\
ORCID: \texttt{0000-0002-1825-4290}} }

\maketitle

\begin{abstract}
The introduction of artificial intelligence into activities traditionally
carried out by human beings produces brutal changes. This is not without
consequences for human values. This paper is about designing and implementing
models of ethical behaviors in AI-based systems, and more specifically it
presents a methodology for designing systems that take ethical aspects into
account at an early stage while finding an innovative solution to prevent human
values from being affected. Two case studies where AI-based innovations
complement economic and social proposals with this methodology are presented:
one in the field of culture and operated by a private company, the other in the
field of scientific research and supported by a state organization.
\end{abstract}

\begin{IEEEkeywords}
artificial intelligence, ethics, value, ethics-by-design, AI applications
% component, formatting, style, styling, insert
\end{IEEEkeywords}

\section{Introduction}

Artificial intelligence (or ``AI'') is often defined as the simulation on a
machine of so-called ``intelligent'' processes. Being both an applied and
theoretical field, this discipline of computer science is covering the range
from weak AI (can machines act intelligently?) to strong AI (can machines
really think?)
\cite{Russell_Norvig__Artificial_Intelligence_A_Modern_Approach__2016}. For the
past decade, the first approach  has brought AI back to the forefront,
especially with the development of new machine learning techniques, such as
deep learning models \cite{Sejnowski__The_Deep_Learning_Revolution__2018}. This
technique has created extremely effective AI applications in the area of
pattern recognition or information selection problems for decision making, with
programs that extract information from raw data and learn to improve their
skills from existing examples. With this learning process, AI systems can
perform complex tasks in place of humans.

However, the arrival of this new technique has brought a number of ethical
issues. Firstly, artificial intelligence programs reason in a simplistic way,
but the real world is complex and full of unexpected events, which the machine
has great difficulty in dealing with. Secondly, when an AI program learns about
data  collected on past situations, it performs statistical deductions and
transforms correlations between variables into implication relationships. This
can lead to problems with dramatic consequences such as gender bias in resume
analysis support system by denying women access to managerial positions
\cite{Dastin__Amazon_AI__2018}, or racial bias in legal decision support system
to predict future criminals \cite{Angwin_et_al__Machine_Bias__2016}.

The concern for the developers is to optimize some specific criteria, e.g.,
efficiency and usability. For instance, online e-commerce marketplaces seek to
optimize the relevance of the connection between supply and demand. They
collect a maximum of information from consumers and suppliers, and use a whole
set of AI-derived strategies to maximize the matching between the different
parties. Attention-economy-based companies seek to keep their users on their
websites and apps for as long as possible to offer them advertising. The
collection of personal data leads to personalization which may have certain
advantages, such as recommending more relevant content, but also important
problems relating to the preservation of privacy, attacks on democracy
\cite{Wylie__Mindfuck__2019}, or the limitation of access to diversity
\cite{Pariser__The_Filter_Bubble__2011}. Intelligent system algorithms are
black boxes that are impossible to understand, they are unregulated and
difficult to question in the case of the presence of bias, and in some cases
they amplify inequalities \cite{ONeil__Weapons_of_Math_Destruction__2016}.

The highlighting of these problems has led to reactions from States and civil
society. Some initiatives, for example in the USA (\textit{Asilomar AI
principles},\footnote{\url{https://futureoflife.org/ai-principles/}} ``AI Now''
Institute...), in Canada (\textit{Montr\'{e}al Declaration for a responsible
development of Artificial
Intelligence}\footnote{\url{https://www.montrealdeclaration-responsibleai.com/}}),
or in France (Villani Report ``For a meaningful Artificial
Intelligence''\footnote{\url{https://www.aiforhumanity.fr/en/}}), with the
production of reports, charters, lists of principles or laws, are moving in the
direction of taking ethical concerns into account early in the development of
AI systems. However, there is not much information provided on how to make this
possible.

\begin{comment}
For designing and implementing models of ethical behaviors, this paper proposes
in the following a methodology for  building ethics-by-design AI systems. This
methodology is explained and illustrated by means of two examples of innovative
applications carried out during PhD studies conducted under the supervision of
the author and how these AI applications are able to respond to problems of
human value conflicts.
\end{comment}

For designing and implementing models of ethical behaviors, this paper proposes
in the following a methodology for building ethics-by-design AI systems. This
methodology is explained and illustrated by means of two innovative
applications developed during PhD works supervised by the author.  We will
explain how these AI applications are able to respond to the problem of human
value conflicts.

\begin{comment}

In the following, for designing and implementing models of ethical behaviors, I
propose a methodology for  building ethics-by-design AI systems. To show how
this methodology works, I will present two examples of innovative intelligent
system applications carried out during PhD thesis studies carried out under my
supervision and how they are able to respond to problems of human value
conflicts.

Human values: what is important to people in their lives, with a focus on
ethics and morality.

As descriptive, ethics attempts (a) to present a typology of what all human
communities experience, in varying degrees and with varying content, as the
objects of moral conscience, that is, the values themselves and their order of
relative worth. It must also (b) describe the elements of moral consciousness,
specifically the noetic structure whereby human beings become aware of values
and make them function in their moral judgments.

As normative, ethics aspires (a) to a critique the various ethea of cultures,
specifically their standards of right and wrong, and to the establishment of a
theory of values that permits a correct evaluation of its norms. Further, it
seeks (b) to characterize the nature of virtue and the contents of the virtues,
and to establish a rank of relative value among them. Thus in its two
functions, ethics must make explicit the elements of moral consciousness and
its objects, and must establish a coherent order among those norms as criteria
of right behavior (as in systems of law or moral teachings).

\end{comment}

\section{Methodology for Ethics-by-Design AI} \label{sec:methodo}

The main idea of the three-step methodology proposed here is to take advantage
of collective intelligence. The introduction of technologies derived from
artificial intelligence leads to risks of undermining values for which the
different stakeholders may have contrasting views.

\subsection{First Step: Innovations in an Economic and Social World}%Innovative Solutions for Maintaining Threatened Values}

The first step of the methodology is to set up an economic and social model
respectful of human values where the contribution of AI technologies can be
integrated as a complement to the skills of human beings. The quote ``They who
can give up essential Liberty to obtain a little temporary Safety, deserve
neither Liberty nor Safety'' attributed to Benjamin Franklin makes us think
that certain values are necessarily contradictory: liberty/privacy vs.
security/safety. \begin{comment}Even though information and communication
technologies (ICT) are often seen as privacy-invading technologies, privacy by
design approaches and technologies try to guarantee this fundamental human
right \cite{Klitou__Privacy_by_Design__2014}.\end{comment} When resources are
limited, choices need to be made. These choices will necessarily follow certain
values and favor certain behaviors supported by some groups of individuals over
others.

However, in the case of adding a new technology such as AI, we must find a way
not to optimize a single value at the expense of others. It is important to try
always to find an alternative and innovative solution that guarantees respect
for all human values. For example, in our study made in the field of predictive
judicial analytics \cite{MuhlenbachSayn2019}, we came to the conclusion that
the values of knowledge and trust are essential to give credit to decisions
produced by intelligent systems. If the deep learning models cannot guarantee
understanding by specialists (i.e., lawyers), they must be replaced by other
more comprehensible models (rule-based approaches, Bayesian methods), otherwise
these AI models will no longer be able to contribute to their role as a tool to
serve justice and help defend litigants.

\subsection{Second Step: Ethical Matrix}

The methodology continues with the clarification of the main values involved in
the introduction of a new AI system for all stakeholders by setting up a
decision-making tool called the ``ethical matrix.'' This tool was initially
proposed by Ben Mepham to facilitate judgements on bioethic questions
\cite{Mepham__2000}, in order to identify the values that are threatened.
\begin{comment} (e.g., the use of genetically modified
fishes, policy interventions in the obesity crisis, or animal sentience
concerns)\end{comment} By focusing more specifically on a small number of
ethical principles (e.g., respect for wellbeing, for autonomy and for justice)
on a given subject (e.g., the impact of new technologies in food and
agriculture), it is possible to elicit the problems and concerns of different
stakeholders or interest groups. The selected ethical principles constitute
values that form the columns of the matrix. The rows of the matrix consist of
each stakeholder caught up with the issue in question. Each cell of the matrix
specifies the main criterion to be satisfied for a stakeholder for a given
principle.
\begin{comment}So, the structure of the ethical matrix consists of stakeholders
on the y-axis, principles on the x-axis, and questions and answers in their
intersections. The ethical matrix can then be seen as a checklist of concerns,
structured around established ethical theory, but it can also be used as a
means of provoking structured discussion between the different interest
groups.\end{comment}

One of the most important difficulties of the ethical matrix approach concerns
the choice of the values to use.  A value is a way of being and acting for a
person or a group of people, and people consider that certain behaviors
resulting from a value are more desirable than others. From Antiquity (e.g.,
with Plato) to the present day, authors have proposed lists of fundamental
values, such as the \textit{Axiology} --defined as the philosophical science of
values-- proposed by the philosopher Max Scheler
\cite{Scheler__Ethics_of_Values__1973}, or the ``Theory of Basic Human Values''
proposed by the social psychologist Shalom~H. Schwartz \cite{Schwartz__1992}.
The arrival of artificial intelligence highlights a specific value: the
\textit{efficiency}. By taking into account a greater number of variables on a
greater number of examples that no human being is capable of accumulating in a
lifetime, intelligent programs are able to accomplish very complex tasks
brilliantly and effortlessly. \begin{comment}Reenacting a new combat between
man and machine, AI is considered as a new avatar of these world-changing
inventions such as the loom, the steam engine or other productivity-improving
technologies, generating as many fears as hopes like any
revolution.\end{comment} However, this efficiency, introduced in a disruptive
manner in fields which until now felt protected, such as medicine or justice,
seems to be predatory for other values.

Faced with this difficult task, it seems important to us to select a list of
basic values, for example the 10-item list of values selected by the
\textit{Montr\'{e}al Declaration for a Responsible Development of Artificial
Intelligence} \cite{MontrealDeclaration__2018}: well-being, respect for
autonomy, protection of privacy and intimacy, solidarity, democratic
participation, equity, diversity inclusion, prudence, responsibility, and
sustainable development. However this list is neither necessary nor exhaustive.
Depending on the application context, values more likely than others to be
affected by an AI technology must replace certain values among the 10 or be
added to this list.

The values of individuals can differ greatly from one individual to another, or
from one social group to another. It is then recommended to take into account
all possible stakeholders, even those who cannot express themselves (e.g.,
animals, nature). In the field of artificial intelligence, the stakeholders
refers especially to researchers, developers, manufacturers, providers,
policymakers, and users \cite{Dignum2019}. In the generic ethical matrix manual
\cite{Mepham_et_al__Ethical_Matrix_Manual__2006}, a protocol has been proposed
for these stakeholders to meet in workshops in order to take into account the
diversity of opinions.

\subsection{Third Step: Value Sensitive Design Approach}

In the last step, the plural visions and their consequences on some human
values provided by the matrix allow to draw the guidelines for integrating
human values throughout the design process and this development by following
the principles of the ``Value Sensitive Design'' (VSD)  introduced by Batya
Friedman and her colleagues in the late 1980s
\cite{Friedman_Hendry__Value_Sensitive_Design__2019}. The VSD theory has
influenced the Dutch approach to responsible innovation
\cite{Handbook_of_Ethics_Values_and_Technological_Design__2015}. In this
approach, human values are integrated throughout the design process and this
development is done using a three-phase survey: conceptual, empirical and
technological \cite{Friedman_et_al__VSD_and_IS__2013}.

There are many different methods of applying this theoretical approach in
practice to real-world problems \cite{FriedmanHB17}. Among the advice given by
the authors, they recommend starting with a value, a technology or a context of
use. Then they propose to identify the direct and indirect stakeholders, as
well as the benefits and harms for each stakeholder group. The benefits and
harms are then mapped onto the corresponding values, allowing the
identification of potential value conflicts.

\begin{comment}

The VSD theory has influenced the Dutch approach to responsible innovation and
value sensitive design. In
\cite{Handbook_of_Ethics_Values_and_Technological_Design__2015}, some examples
are given of design for the values of accountability and transparency, design
for the values of democracy and justice, human well-being, inclusiveness,
presence, privacy, regulation, responsibility, safety, sustainability, and
trust in various domains (alphabetically, from  ``agricultural biotechnology''
to ``water management''). \end{comment}

\section{Case Study 1: Music Recommender Systems}

%\subsection{Digital Music and the Loss of Diversity}

With the advent of the Internet, it was predicted that this would be an
opportunity for musical creation aimed at a niche audience. Unfortunately, it
did not happen.
\begin{comment}The public was going to be able to have access to a greater
diversity of production by musical artists, and these, even if they offered
music of interest only to a niche audience, were going to be able to more
easily find their audience.\end{comment} Apart from a few notable exceptions,
access to music via the Internet has only strengthened trends: popular music,
artists or genres have become even more popular, and lesser-known artists have
found it even more difficult to find their audience, producing an
impoverishment of musical diversity
\cite{Coelho_Mendes__Digital_music__2019}.\begin{comment}: worldwide, according
to the age group, everyone listens to roughly the same thing, with a little
variety depending on their birth tongue and native culture. Yet we have varied
musical desires and changing and evolving musical tastes, with a certain
appetite for musical discoveries.\end{comment}

One of the reasons for the lack of diversity is linked to what Eli Pariser
calls the ``filter bubble'' \cite{Pariser__The_Filter_Bubble__2011}.
Personalized recommendations are based on the processing of big data: the
algorithms recommend to individuals products that are similar, or products
appreciated by individuals sharing common traits, or popular products. With
machine learning, recommendation algorithms will identify people's profiles,
extract the most prototypical traits, and make only recommendations
corresponding to a caricature of everyone's interests. The modus operandi of
the recommender systems may be suitable for mainstream consumers who want to go
directly to the greatest hits. However, this ``superstar effect'' --where the
winner takes all \cite{Coelho_Mendes__Digital_music__2019}-- impoverishes
musical diversity and does not allow musical niches to emerge, ending up
stifling a musical creativity that can no longer reconnect with its audience.

If the digital music market is held only by major record labels supporting a
musical offer limited to a few superstars, are the music and musical artists
produced by independent record labels destined to disappear? In addition, we
have become accustomed to paying the container (e.g., a subscription to an
Internet service provider, a nice smartphone) but not the cultural content as
such (e.g., music and videos on online video-sharing platform, mobile
applications on app stores). As the new digital use is mainly free, how can we
make people accept to pay to access to unknown musical content produced by
indie labels?

According to the methodology proposed here, the \textbf{first step} consists of
a response from the economic and social world. Such a response exists in the
form of a company offering the general public innovative solutions to discover
independent creations. The French social economy company
\textit{1Dlab}\footnote{1Dlab  --- Innovation, Culture \& Territories:
\url{http://en.1d-lab.eu/}} has decided to take up this challenge. To be able
to give an answer to the question of how to support the varied musical creation
produced by independent record labels, this company has found a two-level
solution.

The first level solution is economic. The company proposes a fairer
remuneration model to the creators (artists and content producers). The
business plan works according to a business-to-many model (B2M). The company
provides a music streaming platform (\textit{1DTouch}, \textit{diMusic}) but
also works in the physical world to build a network of places and actors for
favoring new ways of sharing, meeting, discovering and creating a cultural
diversity. The streaming platform access is possible by a subscription fee paid
by the project located partners (libraries, showrooms, work's councils,
administrative divisions...) to their members. \begin{comment}The second level
solution is provided by the work of Pierre-Ren\'{e} Lh\'{e}risson during his doctoral
studies conducted under the supervision of Pierre Maret and myself: it is a
technical innovation allowing the general public to discover pleasant
independent creations with bold recommendation tools that we have designed
following an ethics-by-design approach\end{comment}The second level solution
was proposed in the PhD work of Pierre-Ren\'{e} Lh\'{e}risson, supervised by Pierre
Maret and the author of this paper: it is a technical innovation allowing the
general public to discover pleasant independent creations with bold
recommendation tools that has been designed following an ethics-by-design
approach \cite{Lherisson_et_al_ADMA2017}.

The views on values such as \textit{music} (or more generally to
\textit{culture} as a whole), \textit{equity}, \textit{diversity},
\textit{trust} and \textit{usefulness} are very different depending on whether
the stakeholders are the consumers (who listen to music), the creators of
cultural contents (artists), the media services providers or the independent
record labels.  Following the \textbf{second step} of our methodology, we build
an ethical matrix (Table~\ref{tab:EthicalMatrixDiMusic}) allowing to see how to
develop an innovative solution capable of meeting all the criteria.

%\footnotesize %\small
\begin{table*}
  \caption{Ethical Matrix of AI-based Model of Fair Music Recommender System in an Indie
Streaming Platform.}
  \label{tab:EthicalMatrixDiMusic}
  \begin{center}
  \scalebox{0.85}{\begin{tabular}{|>{\em}c||c|c|c|c|c|c|}
    \toprule
    Respect for: & \emph{Culture (music)} & \emph{Equity} & \emph{Diversity} &  \emph{Trust} &  \emph{Usefulness}   \\
    \midrule
    \midrule
    \makecell{Consumers \\ (Listeners)} &
    \makecell{Have access \\ to a maximum \\ of music} &
    \makecell{Have the same  chance  \\ to  access interesting  music \\ and  make  new  \\  musical discoveries} &
     \makecell{Have access to a \\ maximum of diverse \\ music  genres \\ and music artists} &
     \makecell{Receive music  recom- \\ mendations  adapted to \\ musical tastes and desires} &
     \makecell{Access interesting  \\ and varied music  at \\ a lower cost with  \\ personalized recommendations} \\
    \midrule
    \makecell{Creators of \\  Cultural \\ Goods (Artists)} &
    \makecell{Have access to a maximum \\ of  potentially interesting \\ cultural sources} &
     \makecell{Have the same chance \\ of being recommended, \\ regardless of popularity} &
      \makecell{Be recommended to \\ various  listeners  likely  to \\ appreciate their creations} &
      \makecell{Believe in  the fairness \\ of the  music recommendations} &
      \makecell{Be paid in order to \\  make a living from \\ their creations} \\
    \midrule
     \makecell{Media \\ Services \\ Provider} &
     \makecell{Have the richest  \\ and most varied \\ music catalog} &
    \makecell{Propose  recom- \\ mendations so that \\ all artists can find \\ their audiences} &
    \makecell{Propose a  varied offer \\ intended for specialized \\ and amateur  listeners } &
    \makecell{Have professional  quality \\ music to offer   customers  \\ or subscribers} &
    \makecell{Offer a  user-friendly music \\ platform with an inexpensive \\ solution for customers \\ but allowing artists to  \\ make a living  from their work}  \\
    \midrule
    \makecell{Indie \\ Labels} &
    \makecell{Discover new talents; \\ Support artists' profes- \\ sional development paths; \\ Facilitate  meeting \\ with the audience} &
    \makecell{Give all \\ artists the \\  same chances \\ of success} &
    \makecell{Not concerned \\ (an indie label is  \\ often  specialized in a \\ particular musical genre) } &
    \makecell{Promote quality  music; \\ See all of their \\ artists appear in \\ the recommendations } &
    \makecell{Introduce a  potentially \\ interested public  to the \\ label or a new music genre} \\
    % \\ (e.g., study of \\ judicial processes \\ of people or \\ companies)
% , study of the differences of judgments according to the judges
    \bottomrule
    \end{tabular}
    }
    \end{center}
\end{table*}

\normalsize

In the \textbf{third step} of the methodology, we follow an approach inspired
by the VSD theory to integrate early in the design the values defined in the
previous step for the development of a fair tool for recommending indie
content.

\begin{figure}[htbp]
\centerline{\includegraphics[width=.98\linewidth]{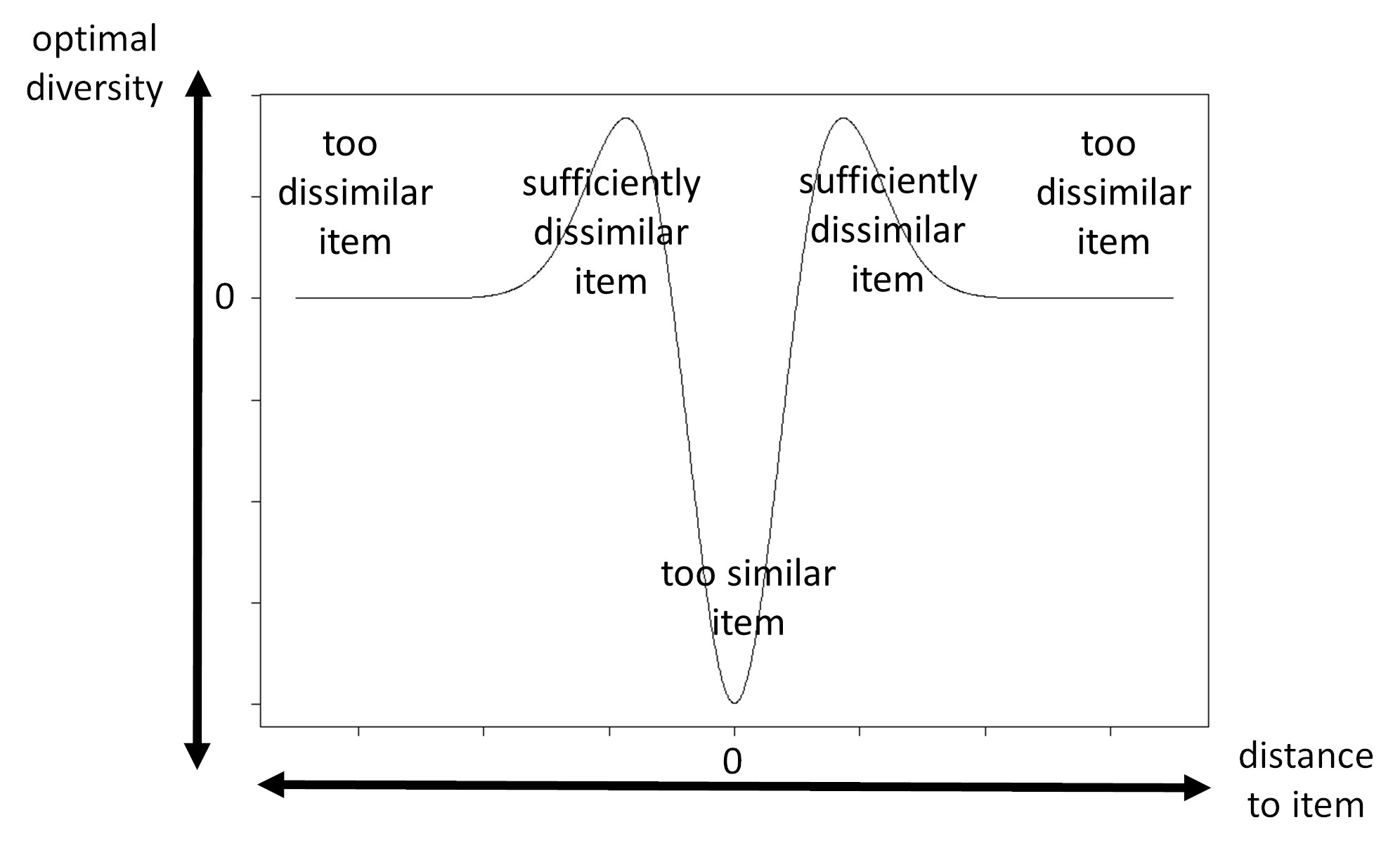}}
\caption{Mexican hat and optimal diversity function. The optimal diversity is
obtained when the items are not perceived too similar nor too dissimilar.}
\label{fig:mexhat}
\end{figure}

First of all, we developed a distance measure between musical items (music,
songs, artists, music genres) based on different criteria (text information of
the music, properties extracted from the sound signal, musical classification
performed by experts) as well as on the perceptual distance sensation obtained
through an experiment with listeners\begin{comment}
\cite{Lherisson__appli_mobile__2017}\end{comment}. Then we proposed an
``optimal diversity function'' using this multi-criteria distance measure in
order to find a compromise between the values \textit{diversity} and
\textit{efficiency} (or \textit{usefulness}).
\begin{comment}To establish this function, we were inspired by the Mexican hat
wavelet, a function which is calculated quite simply because it is the negative
normalized second derivative of a Gaussian function. As an example of use, let
us mention the computation of the connection strength in a self-organizing map
(SOM), an artificial neural network model introduced by Teuvo Kohonen in the
1980s \cite{Kohonen__SOM__2001}. For the SOM, the Mexican hat is used to
amplify the contrasts for neurons distributed in networks of a single layer on
a flat map and the lateral connections are used to create a competition between
neurons. The lateral connections are used to produce excitatory or inhibitory
effects, depending on the distance to a given neuron. The Mexican hat function
describes synaptic weights between neurons in the Kohonen layer: the higher
value of the connection is close to the neuron (the top of the sombrero),
acting as an excitatory effect, then the values decrease as it moves away from
the neuron until they become negative (playing an inhibitory effect).
\end{comment} To get this optimal diversity function, we projected the distances
between items on the (negative) Mexican hat wavelet function to establish clear
distinction between similar, dissimilar, and sufficiently dissimilar items
(Figure~\ref{fig:mexhat}).\footnote{Patent FR3046269,
\href{https://bases-brevets.inpi.fr/fr/document/FR3046269.html}{\textsc{inpi}},
June 2017}

\begin{comment}
is negative for values too close to the item (poor diversity), then it becomes
positive when we move away from the item to a maximum value, and decreases
until it becomes zero if the distance to the item is too important.
\end{comment}

\begin{figure}[htbp]
\centerline{\includegraphics[width=.98\linewidth]{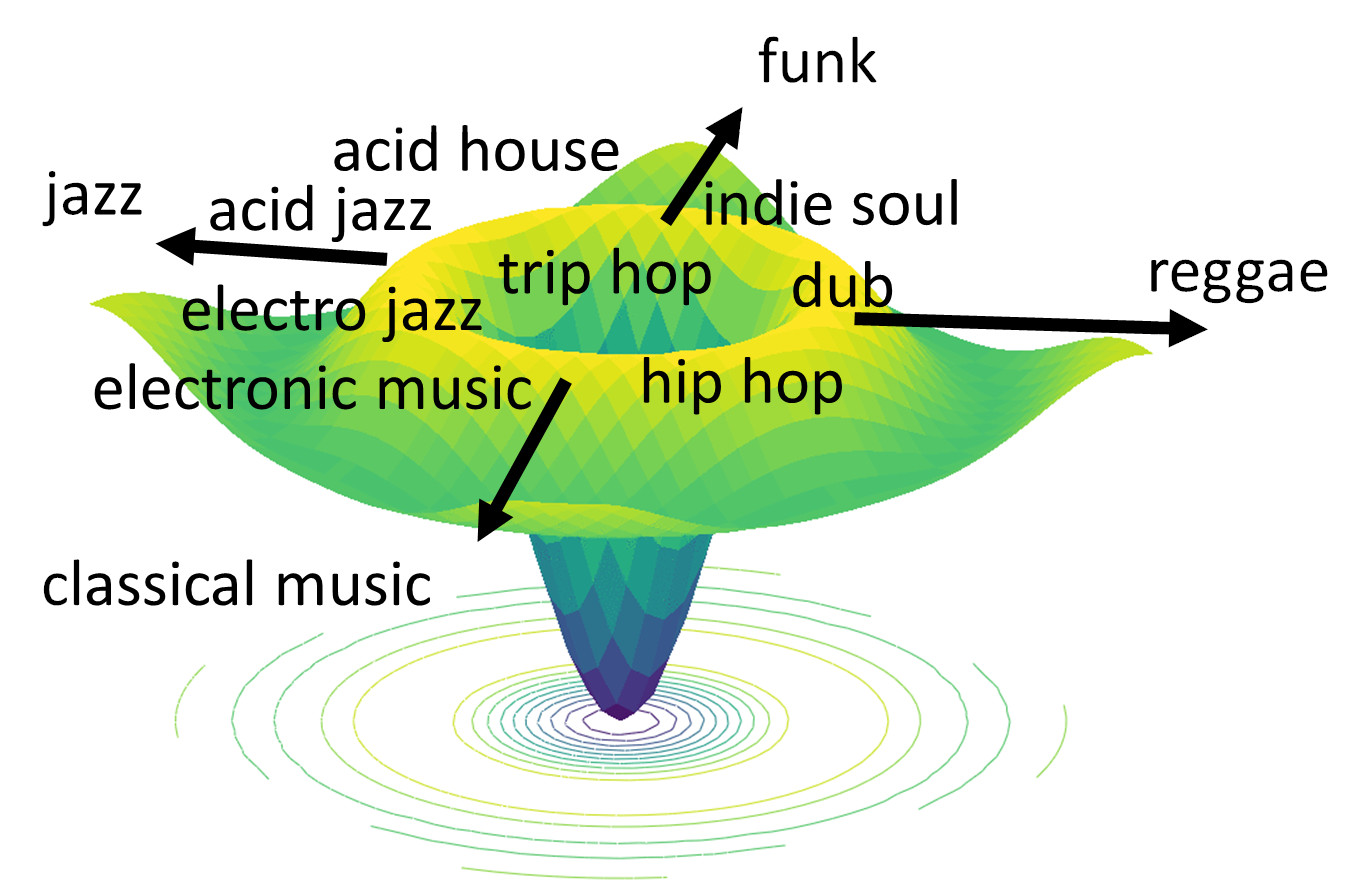}}
\caption{Musical genres and optimal diversity. Compared to a given item (here,
trip hop genre), the diversity will be optimal when it is far enough from the
item in question, but not too much (e.g., acid house, electro jazz, dub or
indie soul can be recommended genres because the diversity is optimal).}
\label{fig:genre}
\end{figure}

For example, a listener enjoys trip hop music, a musical genre resulting from
the fusion of hip hop and electronic music. Based on this information, it is
unreasonable to recommend music located at other ends of the spectrum of
musical genres (e.g., classical music or hard rock). On the contrary, it is
more relevant for the listener to recommend slightly different items, such as
music of the acid house, indie soul or dub type, depending on whether the
musical orientation is respectively rather in the direction of jazz, funk music
or reggae (Figure~\ref{fig:genre}).

Each item (music, song, artist or musical genre) is located in a
multidimensional space where it is always possible to find other items at more
or less close distances, therefore of a more or less optimal diversity value.
In order to guarantee respect for the equity value, it is thus possible to
orient the sense of diversity towards basins of attraction where artists are
less recommended than others in order to make them discovered.

To maintain the trust of the user to the recommendations made by the streaming
platform, the system indicates to the user that the proposed recommendations
are bold, i.e., they are not targeting to the music that the listener is used
to listening to, but that there is a chance that the user may still like it.

If, after having tried these recommended musics in a daring way, the listeners
express their dissatisfaction, the personalization can adapt by decreasing the
distance of optimum diversity and the system can thus propose musics closer to
what they already appreciate. If, on the other hand, the listeners have
eclectic tastes and want bolder discoveries, the distance of optimum diversity
can increase for these people.

Finally, by taking into account the opinions of the various players in a music
platform system at a very early stage, it is possible to propose an innovative
solution guaranteeing respect for a priori contradictory ethical values.

%http://everynoise.com/engenremap.html

% \pagebreak \newpage \clearpage

\section{Case Study 2: Research-paper Recommender Systems in Scientific Digital Libraries}

Access to knowledge is a major problem for human beings. This issue affects
populations of all ages and at all levels: education for all the children of
the planet, fight against (digital) illiteracy, and access to expert knowledge
for all scientists, etc. In the latter case, there are notable initiatives that
are part of an \textit{open science} perspective, but articles published in
scientific journals or renowned conference proceedings still remain mainly the
property of scientific publishers and their access is most often made by
purchasing articles or by subscribing to the publisher's platform. Yet this
access to knowledge is necessary for the production of knowledge because
researchers are science \textit{prosumers}: \begin{comment}when researchers are
not involved in other tasks required by their position,\end{comment}
researchers ``consume'' and ``produce'' knowledge, and this activity takes one
of the two following paths: \textit{exploration} or
\textit{exploitation}\begin{comment}
\cite{Berger-Tal_et_al__The_Exploration-Exploitation_Dilemma__A_Multidisciplinary_Framework__2014}\end{comment}.
In the exploitation phase, researchers use their existing knowledge to produce
new knowledge by following a discovery process specific to their discipline. In
the exploratory phase, researchers update their knowledge by discussing with
colleagues in their laboratory, reading books or scientific articles, or
attending seminars and conferences. Exploratory research is an essential phase
of researcher's activity
\cite{Marchionini__Exploratory_search__From_finding_to_understanding__2006},
and exploitation and exploration are two complementary phases that feed on each
other.

In addition to the cost of accessing these sources of knowledge, even if these
scientific digital platforms are multidisciplinary, the way they are designed
and the means they offer to access interesting information will not promote the
knowledge transfer between disciplines. Each scientific discipline has its own
vocabulary, its own jargon. Without knowing the good keywords to enter in the
search engine of the digital library platform, scientists thus find themselves
isolated in a ``filter bubble'' \cite{Pariser__The_Filter_Bubble__2011}
preventing them from accessing the content of articles from other disciplines.
However, results can be obtained when the research transcends the limits of a
discipline. It is well established that exchanges (knowledge transfers) between
disciplines are very fruitful \cite{Langley_et_al__Scientific_Discovery__1987}.

Faced with these problems, \begin{comment}the Scientific and Technical
Information Department (DIST) of \end{comment}the French state research
organization (CNRS) has proposed a solution that is part of the social and
economic world --as the \textbf{first step} in our methodology-- by launching
an ambitious project in two parts. The first part concerns the creation of a
scientific digital library accessible to all French research institutions
called \textit{ISTEX}\footnote{ISTEX
--- Excellence Initiative in Scientific and Technical Information:
\url{https://www.istex.fr/}}
\cite{CNRS__White_Paper_Open_Science_in_a_Digital_Republic__2016}, resulting
from a massive resource acquisition policy, obtained by entering into operating
contracts with a lot of international and French-speaking publishers.
\begin{comment}The second part concerns the exploitation of these resources in
scientific literature by funding projects making use of techniques applied to
the text, and in particular text and data mining
\cite{CNRS__White_Paper_Open_Science_in_a_Digital_Republic__2017}. During the
PhD studies of Hussein Al-Natsheh that I supervised with Djamel A. Zighed, I
had the opportunity to lead a 4-people team working on such a project with the
following goal: to provide assistance to users of ISTEX resource by offering
recommender systems that promote disciplinary diversity.
\end{comment} The second part concerns the exploitation of these ISTEX resources with text
and data mining techniques
\cite{CNRS__White_Paper_Open_Science_in_a_Digital_Republic__2017}. Together
with the PhD student Hussein Al-Natsheh and with Djamel A. Zighed
(co-supervisor), we had the opportunity to lead a 4-people team working on this
goal: provide assistance to users of ISTEX resource by offering recommender
systems that promote disciplinary diversity.

For researchers (whether they \textit{produce} or \textit{seek} knowledge),
scientific publishers, research organizations and public authorities, interests
diverge greatly, and their respective views on values such as
\textit{knowledge}, \textit{equity}, \textit{diversity}, \textit{trust} or
\textit{usefulness} do not involve the same issues, which can lead to
conflicts, as shown on the ethical matrix (\textbf{second step}:
Table~\ref{tab:EthicalMatrixDigitalLibrary}).

%ETHICAL MATRIX OF AI-BASED MODELS OF RECOMMENDER SYSTEM FOR SCIENTIFIC DIGITAL LIBRARY.

\begin{table*}
  \caption{Ethical Matrix of AI-based Model of Fair Research-Paper Recommender System in a Scientific Digital Library.}
  \label{tab:EthicalMatrixDigitalLibrary}
  \begin{center}
  \scalebox{0.85}{\begin{tabular}{|>{\em}c||c|c|c|c|c|c|}
    \toprule
    Respect for: & \emph{Knowledge} & \emph{Equity} & \emph{Diversity} &  \emph{Trust} &  \emph{Usefulness}   \\
    \midrule
    \midrule
    \makecell{Knowledge-\\ producing \\ Researchers} & \makecell{Want their work to \\ be  accessible to the \\ scientific  community} &
    \makecell{Have the same \\ chance of reaching an \\ interested readership} &
     \makecell{Can be read even by \\ people not from the \\ same scientific discipline} &
     \makecell{Be able to be found \\ when a reader  is \\ interested in their work} &
     \makecell{See their work  \\ appropriately \\ recommended} \\
    \midrule
    \makecell{Knowledge-\\ seeking \\ Researchers} &
    \makecell{Have access to  a \\ maximum of potentially \\ interesting knowledge resources} &
     \makecell{Not to be  limited \\ in  their search \\ for information} &
      \makecell{Have access to \\ various sources \\ of knowledge} &
      \makecell{Believe in the \\ seriousness of the \\ recommended publications} &
      \makecell{Access useful \\ information at low cost} \\
    \midrule
     \makecell{Scientific \\ Publishers} &
     \makecell{Have the most attractive \\ knowledge offer in \\ the  digital  library} &
    \makecell{Not concerned} &
    \makecell{Propose a varied \\ offer intended for \\ specialized communities} &
    \makecell{Have quality  publications \\ to offer to  customers \\ or subscribers} &
    \makecell{Find a  cost \\ effective publishing \\ solution}  \\
    \midrule
    \makecell{Research \\ Organizations and \\ Public Authorities} &
    \makecell{Participate in the increase of \\ knowledge at national level} &
    \makecell{Give all research \\ stakeholders access to \\ information sources } &
    \makecell{Give all  scientific \\ communities the same \\ chances of  being valued} &
    \makecell{Promote trustful and \\ quality research } &
    \makecell{Have an  effective and \\  inexpensive solution } \\
    % \\ (e.g., study of \\ judicial processes \\ of people or \\ companies)
% , study of the differences of judgments according to the judges
    \bottomrule
    \end{tabular}
    }
  \end{center}
\end{table*}

\begin{figure}[htbp]
\centerline{\includegraphics[width=.65\linewidth]{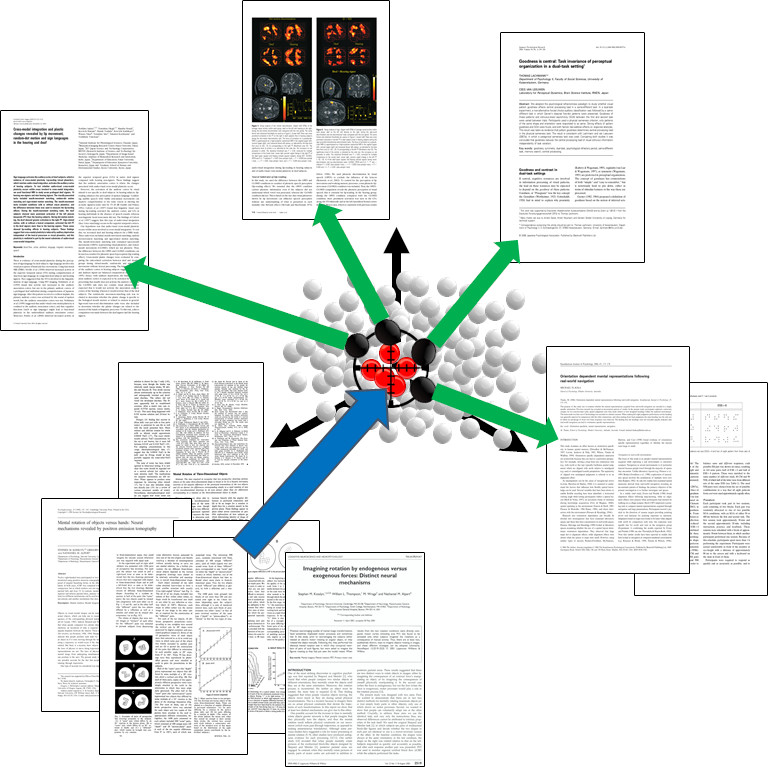}}
\caption{Semantic-similarity based recommendation of scientific papers. First,
a vector representation of the scientific terms is created with word embedding
techniques from the contents of the research articles of the library
(represented by white dots in a multidimensional space). Second, the
interesting papers provided by the user are projected in the vector
representation space (represented by red dots) to define a target area of
interest. Third, the neighboring articles of the target (black dots) in the
vector representation space are activated to recommend diverse items, sometimes
coming from other disciplines, but connected through semantic links to the
input papers.} \label{fig:semsim}
\end{figure}

In the \textbf{third step} of the methodology, we integrate the values of the
ethical matrix into the design of an innovative solution, following the VSD
approach. Here, we focus on the value of diversity which is threatened by the
value of a precision-based usefulness. As part of the projects funded by the
CNRS and intended to benefit from the ISTEX platform, we have proposed to
develop a research-paper recommender system whose purpose is to favor
diversity. Instead of being focused on the sole criterion of accuracy of
results, as most other systems do \cite{Beel_et_al__2016}, we propose a system
able to recommend scientific papers through a semantic similarity model
\cite{Al-Natsheh_et_al__2017} based on computational linguistics methods and
word embedding techniques \cite{Mikolov_et_al__2013}, as shown on
Figure~\ref{fig:semsim}.

In this system, the user does not enter a set of keywords but a list of
articles dealing with a given subject of interest. The content (abstract, plain
text) of these articles is projected in a multidimensional space constructed by
the text representation of the scientific papers of the multidisciplinary
digital library (blue arrow in Figure~\ref{fig:semsim}). As the representation
space is constructed by semantic similarities resulting from a proximity of the
presence of terms in the texts, the more or less distant neighborhood will
bring articles to recommend (arrows in green) comprising more or less diversity
with the input articles. In the experiments we launched, we collaborated with
sports science researchers who were working on a test to detect motor skills.
Thanks to our system, we have been able to recommend research papers to the
sports science researchers --deemed relevant by them-- from different
disciplines such as psychology, linguistics and defectology (e.g., studies on
sign language used by deaf people, another form of complex movements with
meaning). This kind of transfer of knowledge between disciplines has hitherto
never been established.

By examining in detail the different values involved using the ethical matrix,
it was possible to explain them better and to realize that the usefulness of
the system does not necessarily imply the quest for precision: diversity also
matters. The early integration of the diversity value in the design of the
research-paper recommender system allows researchers in the exploratory phase
to have access to knowledge from various sources. \begin{comment}Through the
pooling of resources, access to knowledge is equitable for all. ISTEX platform
allows researchers to be treated fairly: regardless of the financial resources
of the institutions to which they belong, they can access to the same sources
of knowledge.\end{comment} With this recommender system, ISTEX platform
promotes scientific diversity and allows all scientific disciplines to be
represented with equity and all their articles to be recommended. From an
epistemological point of view, this diversity is essential because otherwise a
researcher using a scientific digital library will only have access to sources
of information from his/her own discipline. The result will be that the
documents found will only confirm the initial point of view of the researcher
using this digital library, thus calling into question one of the golden rules
of science which is the property of ``falsifiability''
\cite{Popper__The_logic_of_scientific_discovery__2002}.

\section{Conclusion}

\begin{comment}

In this article, we have seen that the solutions for solving human value
conflicts encountered in the presence of an AI system are never carried out
only by IT specialists. The solution can emerged from the exchange between the
various stakeholders.

innovations based on artificial intelligence techniques appear to be a clever
way of answering questions that have fundamental repercussions on society and
risk undermining certain human values.
 At the time of writing, we are in an unprecedented situation of
home lockdown due to the COVID-19 crisis. The 2019-2020 coronavirus pandemic
has spread all over the Earth, and this major international public health
problem is causing serious socio-economic disruption in the world, including
the greatest global recession in nearly a century.

The world is undeniably changing, and this health-related event can be seen as
a trigger for the extreme reactions of people, with the appearance of behaviors
of great solidarity between individuals and assistance to the most deprived, as
well as the moment when selfishness reactions and feelings of xenophobia are
exacerbated (e.g., against people perceived as being Chinese because they are
seen as the source of the infection). \end{comment}

AI is often associated with robotization, and robotization can be understood as
``automation'' but also as ``the process of turning a human being into a
robot.'' That is why the arrival of AI can scare some individuals, giving rise
to feelings of technophobia with the fear of mass unemployment due to jobs
performed by robots or intelligent computer systems. Moreover, there is a
rejection of AI because this technology may be seen as a kind of automated and
not human --if not ``inhuman''-- thinking. The species of the human being is
\textit{Homo sapiens}, which means ``wise man'' in Latin. The human being is
not simply characterized as being intelligent, that is to say having an ability
to find solutions to complex problems, but also has a nature of a discerning,
wise, and sensible being.

The events that we are currently experiencing --COVID-19 crisis, climate
change-- produce abrupt changes in all areas, bringing new challenges in the
fields of health, economy, ecology, security, justice, science, culture\ldots
With the collective intelligence of human beings, let AI and ICT be the means
pushing us to be innovative in order to respond to these problems in an
appropriate manner, and may this response be done wisely in order to guarantee
respect for human values.

\bibliographystyle{IEEEtran}
\bibliography{FMbib}

\end{document}